\documentclass[acmsmall,nonacm]{acmart}

\usepackage{amsmath}
\usepackage{amsfonts}

\usepackage[utf8]{inputenc}
\usepackage[english]{babel}

\usepackage{graphicx}
\usepackage{caption}
\usepackage{subcaption}

\usepackage[binary-units=true]{siunitx}

\usepackage{cleveref}

\usepackage{glossaries}
\newacronym{cnn}{CNN}{convolutional neural network}
\newacronym{nn}{NN}{neural network}
\newacronym{scn}{SCN}{sparseconvnet}
\newacronym{scl}{SCL}{sparse convolutional layer}
\newacronym{ascl}{ASCL}{asynchronous sparse convolutional layer}

\makeglossaries

\begin{document}
    \title{Transferring dense object detection models to event-based data}
    \author{Vincenz Mechler}
    \email{vincenz.mechler@igd.fraunhofer.de}
    \affiliation{%
     \institution{Fraunhofer IGD}
     \city{Darmstadt}
     \country{Germany}}
  
    \author{Pavel Rojtberg}
    \email{pavel.rojtberg@igd.fraunhofer.de}
    \affiliation{%
     \institution{Fraunhofer IGD}
     \city{Darmstadt}
     \country{Germany}}
      
    \setcopyright{none}
    \acmConference[AIVR '22]{AIVR '22: The 6th International Conference on Artificial Intelligence and Virtual Reality}{July 22--24, 2022}{Kumamoto, Japan}
    \acmBooktitle{AIVR '22: The 6th International Conference on Artificial Intelligence and Virtual Reality, July 22--24, 2022, Kumamoto, Japan}
    \copyrightyear{2022}
    \acmYear{2022}
    
	\begin{abstract}
        \textit{Event-based} image representations are fundamentally different to traditional \textit{dense} images. This poses a challenge to apply current state-of-the-art models for object detection as they are designed for dense images.
        In this work we evaluate the YOLO object detection model on event data. To this end we replace dense-convolution layers by either sparse convolutions or asynchronous sparse convolutions which enables direct processing of event-based images and compare the performance and runtime to feeding event-histograms into dense-convolutions.
        Here, hyper-parameters are shared across all variants to isolate the effect sparse-representation has on detection performance.
        At this, we show that current sparse-convolution implementations cannot translate their theoretical lower computation requirements into an improved runtime. 
    \end{abstract}


\begin{teaserfigure}
    \centering
    \includegraphics[width=0.88\textwidth]{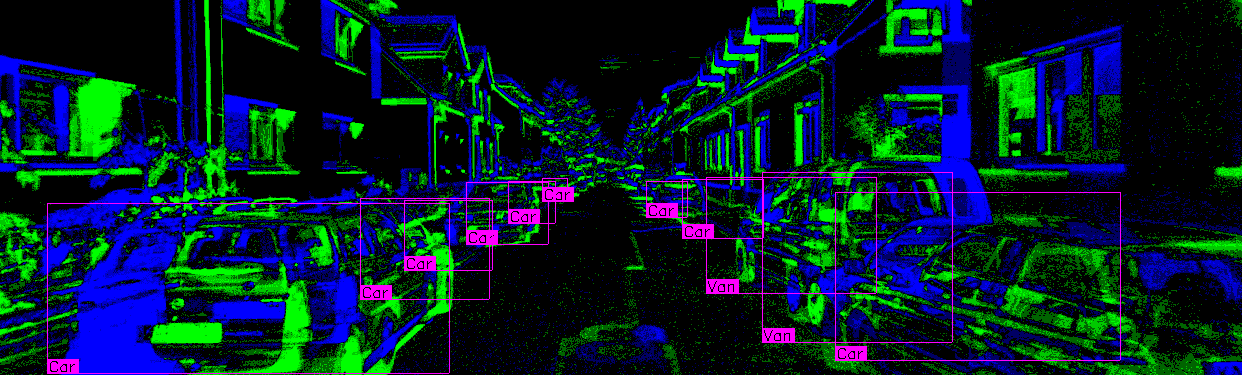}
    \includegraphics[width=0.88\textwidth]{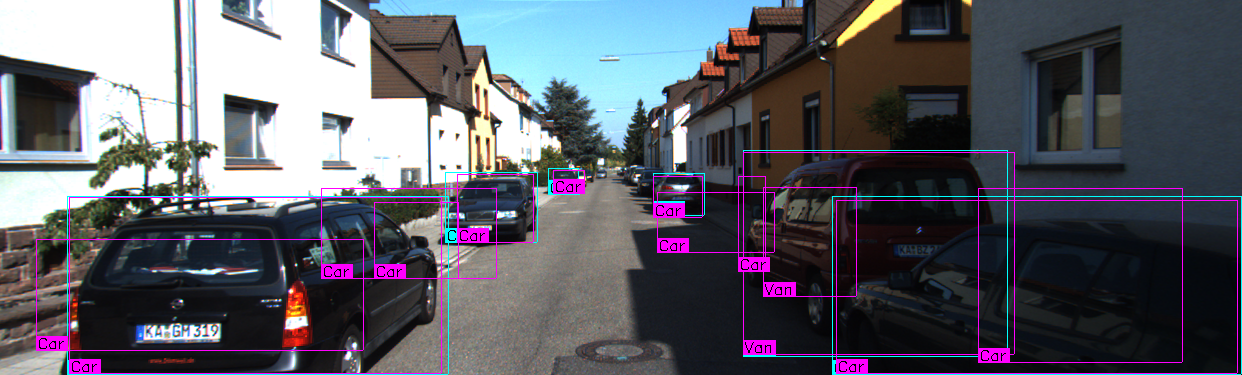}
    \caption{Object detection on the KITTI dataset \cite{kitti}. Cyan boxes denote ground-truth. Pink boxes denote predictions. \textit{Top:} Using sparse event-histograms \textit{Bottom:} Using source RGB-image responsible for the off events. }
\end{teaserfigure}


	\maketitle
	\newpage
	\section{Introduction}
	\label{sec:introduction}
	    Event-based or neuromorphic cameras provide many advantages like high-frequency output, high dynamic-range and a lower power-consumption.
        However, their sensor output is a sparse, asynchronous image-representation, which is fundamentally different to traditional, dense images.
        
        This hinders the use of convolutional layers, which are an essential building-block of current state-of-the-art image processing networks.
        Classical convolutions on sparse data, as it is produced e.g. by event-cameras, are inefficient, as a large part of the computed feature-map defaults to zero. Furthermore, sparsity of the data is quickly lost, as the non-zero sites spread rapidly with each convolution.
        To alleviate this problem, changes to the convolutional layers were proposed.
        
        Sparse convolutional layers \cite{scn} compute convolutions only at active (i.e. non-zero) sites.
	    The sub-type of 'valid' or 'submanifold' sparse convolutional layers furthermore tries to preserve the sparsity of the data by only producing output signals at active sites, which makes them highly efficient at the cost of restricting signal propagation.
        Non-valid sparse convolutions are semantically equivalent with dense convolution layers in that they compute the same result given identical inputs. Valid or submanifold sparse convolution layers, on the other hand, differ from dense convolutions, but still provide a good approximation for full convolutions on sparse data.
       
       	\citet{rpg} further introduce asynchronicity into the network. This allows for samples to be fed into the network in parts as they are produced by a sensor, and thus to reduce the latency in real-time applications.
        Several small batches of events from the same sample can be processed sequentially, producing identical results to synchronous layers once the whole sample has been processed.
        However, \cite{rpg} only implemented a proof-of-concept. The project only includes asynchronous submanifold sparse convolutional and batch-norm layers, whereas the \gls{scn} project\cite{scn} provides a full-fledged library.        
        Furthermore, asynchronous models cannot be trained, as the \textit{index\_add} operation used in the forward function is not supported by PyTorch's automatic gradient tracking. This, however, does not pose a problem, as each layer is functionally equivalent to its \gls{scn} counterpart. Therefore, it is possible to train an architecturally identical \gls{scn} network and transfer the weights. As the asynchronous property is only relevant during inference, this does not pose a limitation.
                
        An alternative approach is to convert the sparse frame representation to dense frames first, using a learning-based approach \cite{e2vid}. This way, however, one loses all computational advantages that the sparse representation offers.
        Notably, it is also possible to synthesize events from a dense frame-based representation \cite{vid2e}.
	
        Furthermore, a \textit{leaky surface layer} was proposed by \citet{yole} which integrates the event-to-frame conversion directly into the target network. This way the network becomes stateful, and resembles a spiking model \cite{snn}.
        
	\subsection{Contributions and outline}
	In this work, we use the YOLO v1 model \cite{yolo-paper} as a simple but powerful dense object recognition baseline. We model sparse networks architecturally identical to YOLO v1 using the \gls{scn}\cite{scn} and asynet\cite{rpg} frameworks. These serve as a case study to evaluate the performance of sparse and asynchronous vs dense object detection.

	We implement all variants in PyTorch and evaluate the predictive performance and runtime requirements against a dense variant. To this end, we convert the KITTI Vision dataset to events using \cite{vid2e}. This allows us to answer the question if these novel technologies are a viable optimization over dense convolutional layers, or if they fall short of the expectations in practice.

	The remaining part of this work is structured as follows: First, \cref{sec:data} introduces data formats required for the remainder of this work. Next, \cref{sec:implementation} details the major changes and additions to the used frameworks. \Cref{sec:evaluation} evaluates the sparse and dense YOLO versions w.r.t. performance, and \cref{sec:profile} regarding runtime. \Cref{sec:conclusion} concludes our work by discussing our results and providing an outlook.
	
	\section{Dataset and frame representation}
	\label{sec:data}
	
    In this work we focus on frame-based object detection models. Dense video-frames capture absolute light intensity of a scene at a specific moment in time.
	Event cameras, on the other hand, capture discrete changes of brightness at specific locations in space and time as a sequence of events. These are more akin to videos, or image sequences, but without knowledge of the absolute brightness of any location.
	
	Synthetic event sequences can be generated from videos by interpolating to \textit{infinite} resolution in the time axis and then extracting each change of a pixels value as a discrete event at the (estimated) time of this change. Vice versa, videos can be reconstructed from events, if the absolute information of the \textit{starting frame} is known. Without such information, one can still try to reconstruct a single image from events, e.g. by assuming the starting point to be an empty, grey image, which might yield good results if the amount of events available is large enough.
	
	\subsection{KITTI Vision dataset}
	
	As we compare sparse with dense \glspl{cnn}, we require parallel dense image and sparse event data. Therefore, we use the dense KITTI Vision object detection dataset \cite{kitti} and convert it to events using vid2e \cite{vid2e}.
	
	The dataset contains a large collection of urban traffic based scenes captured by cameras mounted on the roof of a car. Each sample is manually annotated with bounding boxes of different object classes like 'Pedestrian', 'Car', 'Cyclist', etc. The intended use is autonomous driving and driver-assistance systems.
	
	\subsection{Optical event data and histograms}
	
	Optical event data can be represented in various formats. A simple and lossless encoding are sequences of discrete events storing the spatial and temporal location and the polarity of the change  \cite{event-vision}, as the amount is usually assumed to be fixed within one dataset.
	
	This format is, however, badly suited to processing with e.g. \glspl{cnn}, as the sequence length is variable across samples and unbounded. A common format that overcomes this limitation are event-histograms, which accumulate all events into a single frame similar to an image, but showing changes of brightness during the defined interval instead of absolute brightness values at a single point in time.
	
	In this work, we use the event-histogram representation from the asynet \cite{rpg} framework producing two channels, where each pixel value represents the sum of all observed event changes of negative or positive polarity at this spatial location.
	
	\section{Implementation}
	\label{sec:implementation}
	In addition to the available dense PyTorch YOLO v1 implementation\footnote{https://github.com/zzzheng/pytorch-yolo-v1}, we implemented two more networks to be used in our final evaluation: A sparse version of YOLO v1 implemented in the \gls{scn} framework and an asynchronous sparse version implemented in the asynet framework.
    Training and evaluation of all networks is performed using the asynet framework to ensure comparability of results.
    For reproducibility, we make our implementation available as open-source on github\footnote{https://github.com/paroj/rpg\_asynet}.
	
	The sparse models follow the YOLO v1 architecture, but use specialized sparse or asynchronous sparse layers in the convolutional block, followed by standard PyTorch linear layers.
    In the \gls{scn} sparse model, the convolutional block is followed by a sparse-to-dense layer that converts the sparse tensor into a dense representation for further processing.
    In the asynet asynchronous model such a layer is not necessary, as the model does not support training anyway and the dense feature map tensor is passed through the network alongside the sparse events as part of the sparse representation.
	
	Both sparse models employ submanifold sparse convolution layers where the dense network uses convolutions with stride $1$ to achieve maximum performance.
	We adapted the trainers for dense and sparse object detection models already implemented in the asynet code for improved logging and debugging and added early stopping.
	However, neither the \gls{scn}, nor the asynet framework contained sufficient functionality to directly implement a YOLO v1 network.
	
	\subsection{Sparseconvnet extensions}
	\label{sec:same-padding}
	
	In the case of \gls{scn}, the deficit was minimal, as it only lacked the 'same'-padding feature in its sparse convolutional layer. To get around that limitation, we chose a rather inefficient but easy way of converting a sparse tensor into a standard dense PyTorch tensor, pad this dense representation, and then convert it back into a SCN sparse tensor.	
	This does not affect our evaluation, as it can be easily excluded from the runtime evaluation carried out via profiling, and does not change the results of the layer computations.
	
	\subsection{Asynet extensions}
	\label{sec:extensions-NonValidSparseConvolution2D}
	
	The asynet framework, however, was missing a layer type. The existing 'asyncSparseConvolution2D' layer implements an asynchronous valid or submanifold sparse convolution only. The project does not contain an implementation of an asynchronous (non-valid) sparse convolution.
	We therefore implemented the \textit{asynNonValidSparseConvolution2D} layer, based off the \textit{asyncSparseConvolution2D} implementation.
	To ensure correctness, we again specified test cases to verify our implementation.
    
    Additionally, the original code did not filter duplicate events within a sequence, causing each active site to be processed as often as the number of duplicate events (at the same spatial location) in the sequence. This behaviour caused the runtime to increase by several orders of magnitude, while also producing incorrect results in case of duplicate events.
    
	\subsection{Dataloader}
	
	We implemented a dataloader for the KITTI Vision dataset analogously to the already available dataloaders for various other datasets (NCaltech101 among others). We adapted code available through the dataset's release site \cite{kitti-web} for converting the ground truth bounding boxes and labels into the commonly used format defined by the Pascal VOC dataset. Each sample is converted to events at runtime because of the enormous storage overhead of preprocessing the whole dataset. The spatial locations of the events are then rescaled to the required tensor size, and finally accumulated into an event histogram.
	Additionally, we implemented a version without the conversion into events to be used to train the dense YOLO network on the original images.
	    
	\section{Error analysis}
	\label{sec:evaluation}
	
	The intuition behind sparse \glspl{cnn} is to speed up, and reduce energy consumption of, dense \glspl{cnn} by eliminating unnecessary computations. As such, we require sparse \glspl{cnn} to match the prediction performance of dense \glspl{cnn}, while reducing resource consumption.
	
	We start by verifying the first condition, namely recognition performance matching that of the dense model.
	Due to the high costs of training image detection networks most evaluations were performed only with limited redundancy, as can be seen in \cref{tab:results}. As the goal of this evaluation is a qualitative comparison of different architectures, rather than trying to achieve state-of-the-art results, it is acceptable to omit hyper-parameter tuning and use the same parameters for all models. A proper convergence of each training run, as well as the absence of strong outliers within the performed experiments, minimizes the risk of non-representative and non-reproducible results.

 	\begin{table}
    \begin{tabular}{ |l|l|l|l||c|c| } 
        \hline
        model & data & window-size & med. mAP & med. loss\\ 
        \hline
        \hline
        dense YOLO & dense images & n/a & 0.1914 & 0.4465 \\ 
        \hline
        & event-histograms & \SI{33}{\milli\second} & 0.1777 & 0.4448 \\ 
        \hline
        & event-histograms & \SI{42}{\milli\second} & 0.2055 & 0.3921 \\ 
        \hline
        sparse YOLO & event-histograms & \SI{8}{\milli\second} & 0.2301 & 0.3410 \\ 
        \hline
        & & \SI{16}{\milli\second} & 0.2332 & \textbf{0.3394} \\ 
        \hline
        & & \SI{25}{\milli\second} & \textbf{0.2337} & 0.3413 \\ 
        \hline
        & & \SI{33}{\milli\second} & 0.2115 & 0.3742 \\  
        \hline
        & & \SI{42}{\milli\second} & 0.2321 & 0.3443 \\ 
        \hline
        & & \SI{50}{\milli\second} & 0.2311 & 0.3436 \\ 
        \hline
    \end{tabular}
    \caption{Median mAP and YOLO loss values over 3 training runs for different models, data, and sparse event-window size. Best values highlighted.}
    \label{tab:results}
    \end{table}

	\subsection{Sparse YOLO vs dense YOLO}
	
	We first compared the dense YOLO network trained on dense images directly with our structurally identical sparse implementation trained on 42ms event-windows.
	
	The mAP score shows the sparse model to perform approximately \SI{20}{\percent} better than the dense baseline. This indicates that both the conversion from dense images to sparse events and our model implementation work as intended. 
	
	The increase in performance can be explained by the availability of additional information: The dense model is restricted to a single image per sample, the sparse model, however, is trained on events synthesized from a sequence of images. While events encode only the change and lose the information about absolute brightness, it can be argued that the change, which effectively encodes moving edges, is more beneficial to object recognition than colour and absolute brightness.
	
	The maximum achieved mAP of about $23\%$ is significantly lower compared to values YOLO v1 reportedly achieved on other datasets. This is likely due to differences of the mAP calculation\footnote{https://github.com/thtrieu/darkflow/issues/957}. However, we use the same calculation throughout this work, which makes our results comparable to each other.
	
	The 'YOLO loss', as presented in the original YOLO paper \cite{yolo-paper}, shows considerably larger differences between the sparse and dense variants.
    This metric, however, constitutes a loss function to be optimized during training and is not necessarily suited to compare the performance of different models.
	
	\subsection{Sparse vs dense convolutions}
	
	Training the dense YOLO network on event histograms instead of dense images did not improve its performance notably. 
	It consequently also performed approximately \SI{15}{\percent} worse than the sparse model trained on the same data. This indicates the submanifold sparse convolutions, which are the only semantic difference from the dense model, essentially contribute to the ability of the model to process sparse event data. 
	
	While similar results might be achieved using only dense convolutions by altering the model structure, it is stunning that submanifold sparse convolutions enable the transfer of a model optimized for dense images to events without requiring additional hyper-parameter tuning.

	\subsection{Event-window size}
	
	Against our expectations, event-window size within a reasonable range has little to no effect on the performance. We tested several configurations, starting from \SI{42}{\milli\second}, which covers exactly one frame of the original dense dataset captured at a frame rate of 24fps. As this frame rate is, however, rather low, and \SI{42}{\milli\second} contains a huge amount of events, we tested mainly values below that.
	
	As each training run takes more than one day on our available hardware, we decided to use fixed, pre-trained weights for initialization instead of averaging over multiple, randomly initialized runs to exclude effects of weight initialization from this evaluation. All sparse test cases were trained from the same weights, pre-trained for 121 epochs using a window size of \SI{33}{\milli\second}.
	There is no noticeable qualitative difference in the evaluation loss and mAP scores between the tested window sizes of \SI{8}{\milli\second}, \SI{16}{\milli\second}, \SI{25}{\milli\second}, \SI{33}{\milli\second}, \SI{42}{\milli\second} and \SI{50}{\milli\second}.

	
	\section{Runtime analysis}
	\label{sec:profile}
	
	Sparse \glspl{cnn} claim to be more efficient in terms of number of operations and subsequently runtime and energy consumption. To verify this claim we profiled the three different implementations using cProfile\footnote{https://docs.python.org/3/library/profile.html}. 
	
	For these experiments we chose to evaluate the full YOLO v1 network on real data. This ensures a realistic ratio of active sites over locations in the tensor, as well as number of events per active site. 
	While in the dense framework all convolution layers are essentially the same layer type, all unstrided sparse convolutions can be translated to more efficient submanifold sparse convolutions, while strided sparse convolutions have to be implemented as non-valid sparse convolution layers.
	
	
	Due to the high runtime of the asynchronous framework, all models are profiled over 1042 fixed samples (\SI{14}{\percent}) of the KITTI Vision dataset.

	\subsection{Dense vs sparse}
	
	Although in theory more efficient on sparse data than dense convolutions, sparse \glspl{cnn} still show higher runtimes. Due to highly optimized code and better hardware support for the massively vectorized operations of the standard dense convolutions implemented in the PyTorch framework, the research oriented proof-of-concept implementations of sparse convolutions, while also highly optimized in the SCN framework, cannot compete in real use-cases yet.
	
\begin{figure}
		\centering
		\begin{subfigure}[t]{0.45\textwidth}
			\centering
			\includegraphics[width=\textwidth]{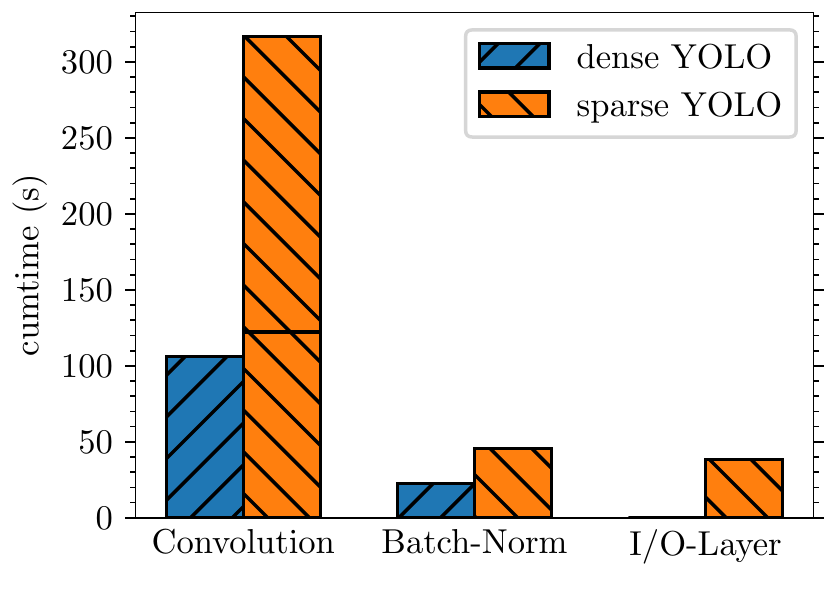}
			\caption{Sparse network shows higher runtime for all layers, but especially for the convolutional layers with approximately triple runtime. Sparse-convolution time is split into submanifold (top) and non-valid (bottom) parts.}
			\label{fig:profile-sparse-dense}
		\end{subfigure}
		\hfill
		\begin{subfigure}[t]{0.45\textwidth}
			\centering
			\includegraphics[width=\textwidth]{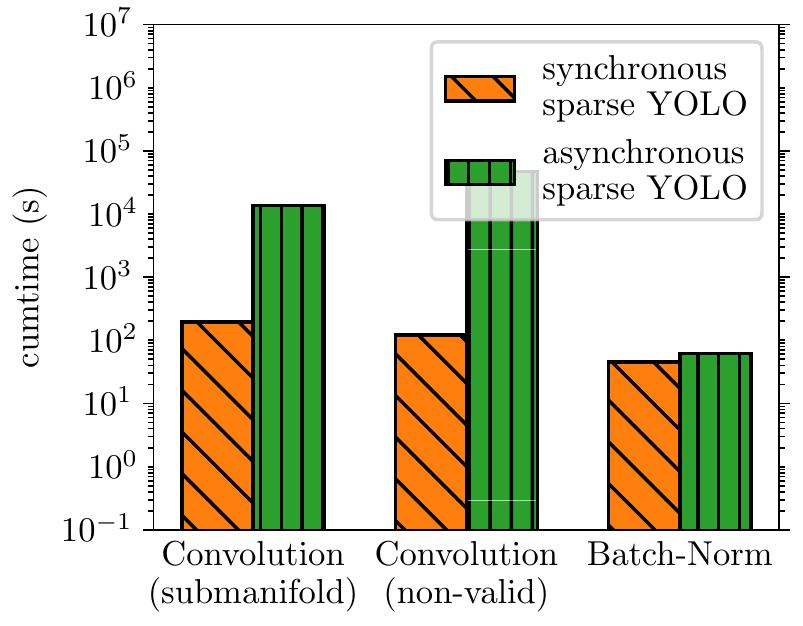}
			\caption{Asynchronous network shows very high runtime (approx. three orders of magnitude higher than the synchronous implementation). \\Y-axis in logarithmic scale due to extreme differences.}
			\label{fig:profile-syn-asyn}
		\end{subfigure}
	\caption{Cumulative runtime (over 1042 samples) of dissimilar layers of dense, sparse, and asynchronous implementations of the YOLO v1 network during prediction.}
	\end{figure}

	Figure \ref{fig:profile-sparse-dense} shows the cumulative runtime of those layers that are not identical in dense and sparse networks. Convolutions in sparse networks are split into (non-valid) sparse-convolutions and submanifold-sparse-convolutions. Here the performance difference is most significant, with an increase in runtime of more than three times. Furthermore, convolutional layers usually constitute the largest part of \glspl{cnn}.
	
	Batch norm layers only experience a small loss in performance. I/O-Layers, which set up additional data structures to be passed through the network with each sample to enable efficient computation of the convolutional layers, are only needed in sparse networks and provide an additional notable overhead. As the YOLO v1 model contains two strided convolutional layers and only one actual input layer, however, about two third of this overhead can be attributed to our implementation of the 'same'-padding option as described in \cref{sec:same-padding}.
	
	\subsection{Batch size}
	
	Repeating this experiment with a smaller batch size of 1 (instead of 30 in the previous evaluation) revealed sparse layers don't suffer as much overhead from smaller batch sizes as dense layers, or rather, in the reverse direction, don't gain as much from predicting more samples at the same time using higher batch sizes.
	
	For convolutions, the gap between dense and sparse layers closes significantly, while sparse batch norm layers actually overtake their dense counterpart. For sparse networks, I/O layers show a similar overhead to convolutions.
	
	\begin{figure}
		\centering
		\begin{subfigure}[t]{0.45\textwidth}
			\centering
			\includegraphics[width=\textwidth]{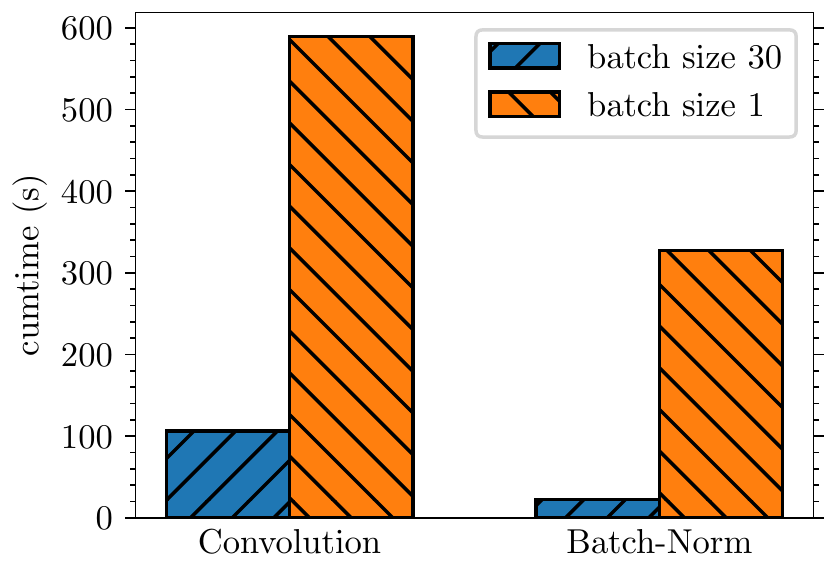}
			\caption{\textit{Dense:} Individual processing results in approximately sixfold increase of runtime.}
			\label{fig:profile-batch-size-dense}
		\end{subfigure}
		\hfill
		\begin{subfigure}[t]{0.45\textwidth}
			\centering
			\includegraphics[width=\textwidth]{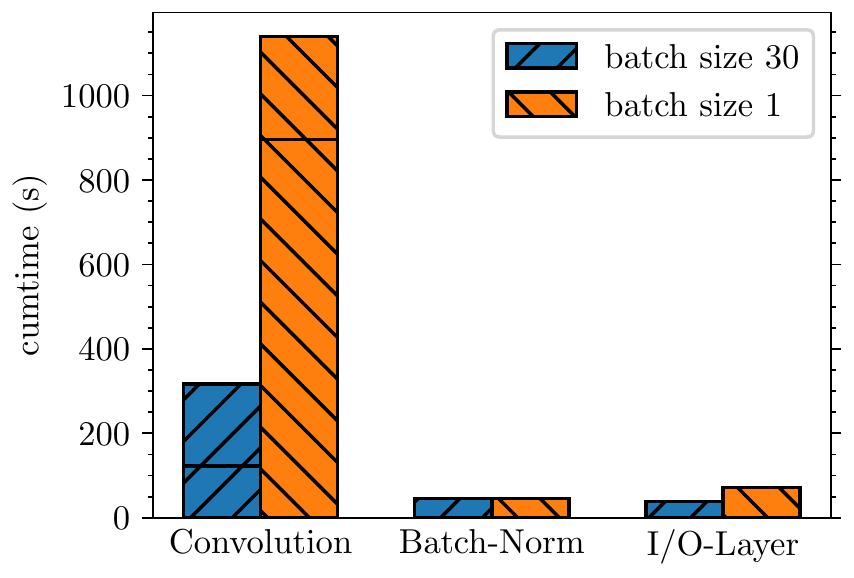}
			\caption{\textit{Sparse:} Individual processing results in approximately threefold increase of runtime. Convolution time is split into submanifold (top) and non-valid (bottom) parts.}
			\label{fig:profile-batch-size-sparse}
		\end{subfigure}
		\caption{Cumulative runtime (over 1042 samples) over batch size for dense and sparse layers.}
		\label{fig:profile-batch-size}
	\end{figure}
	
	\subsection{Synchronous vs asynchronous}
	
	The results of profiling the asynchronous sparse \gls{cnn} implemented in the asynet framework are far from encouraging. Due to the experimental and little-optimized implementation, the asynchronous convolution layers show an increase in runtime of roughly three orders of magnitude, as seen in Figure \ref{fig:profile-syn-asyn}.
    While the synchronous framework essentially works on event histograms accumulating all events at the same location, for the asynchronous framework multiple events per active site are an important feature of the data. An asynchronous network processes a sample in multiple sequences representing discrete time steps. An active site is only processed in any distinct sequence if there exists an event at that active site at a time that falls into the range of that sequence.
    
	\subsection{Asynchronous sequence count}
	
	To make use of the asynchronous nature of the network, a sample will usually be split into a number of event-sequences, which are then processed in series. When examining the effect of the number of theses sequences, we would expect an increase in runtime for larger sequence counts due to possibly duplicated active sites, bounded by the runtime of the 1-sequence-baseline times the number of sequences. For small numbers of sequences the number of active sites in each sequence will not decrease notably, as there usually is more than one event at most active sites in a sample, thus performing close to this upper bound. For large sequence counts, however, we expect a sub-linear increase in runtime due to a decreasing number of active sites per sequence. 
	
\begin{figure}
		\centering
		\begin{subfigure}[t]{0.45\textwidth}
			\centering
			\includegraphics[width=\textwidth]{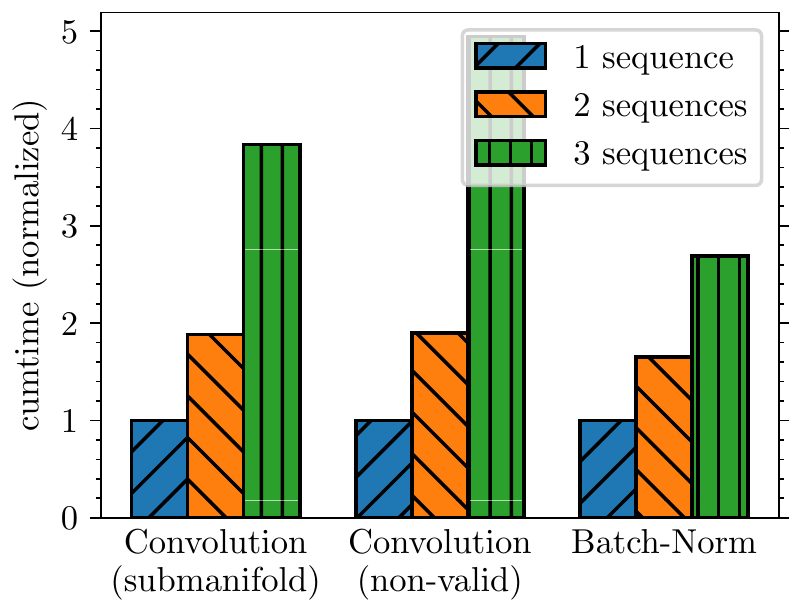}
		\end{subfigure}
		\caption{Normalized cumulative runtime (over 148 samples) of layers of the asynchronous YOLO v1 network at different sequence counts during prediction. Convolution increases super-linearly for sequence counts above two.}
		\label{fig:profile-asyn-sequence-count}
	\end{figure}

	While profiling the model using two sequences exactly matches our expectations, as shown in Figure \ref{fig:profile-asyn-sequence-count}, the model exceeded the upper bound for three sequences. 
	
	Further analysis showed that the unexpected increase in runtime is caused in low-level functions like tensor formatting. We believe this to result from overhead due to extremely high memory utilization. While profiling with three sequences required \SI{360}{\giga\byte} of RAM, profiling of higher sequence counts exceeded our available resources and was thus omitted. Therefore, confirmation of our claim of sub-linear increase in runtime for high sequence counts is left for future work.
	
	
	
	\subsection{Theoretical runtime}

    Given the data is sufficiently sparse, the sparse convolution based method should have a significant advantage.	
\begin{itemize}
    \item 	The dense convolution convolves the filter with every position of the input tensor, yielding as many convolution operations as there are unique locations in the tensor.
    \item 	The sparse convolution convolves the filter with every active site, with the active sites being a subset of the unique locations in the tensor.
\end{itemize}
	Therefore, the complexity of the sparse convolution is generally lower than that of the dense convolution.
	Additionally, submanifold sparse convolutions further reduce complexity by only computing those parts of the convolution at each active site where filter and active sites overlap. The main gain of submanifold sparse convolutions, however, lies in preventing the increase in the number of active sites, additionally reducing the complexity for the following layers.
	
	In practice, sparse convolution layers require the construction of a so-called rulebook for each sample to efficiently compute the necessary convolutions, as detailed in \cite{scn}. While this creates some overhead, it is outweighed by the gains of only processing the active sites. Furthermore, blocks of consecutive submanifold convolution layers pass through the rulebook and ensure it stays valid, so that it only needs to be computed once, further reducing the complexity.
	
	Asynchronous-sparse-convolutions split the events into multiple sequences, and within each sequence behave like synchronous sparse convolutions.
	Each active site of a synchronous-sparse-convolution is caused by at least one event, but might accumulate many events that happened at the same spatial location. 
	Therefore, each active site is processed in at least one sequence, but at worst case in all of them.
	Such layers thus have at least as high a complexity as their synchronous counterparts, but stay within a predictable margin.
    
    However, the sparse nature of the data hinders SIMD processing and the use of on-chip caches --- two techniques that are crucial for reaching high performance on current hardware.
	
	\section{Conclusion}
	\label{sec:conclusion}
	In this work we have evaluated the prediction performance and runtime of sparse and asynchronous-sparse \glspl{cnn} with respect to classical dense \glspl{cnn}. 
	Our experiments have shown that sparse \glspl{cnn} can match the performance of their dense counterparts without requiring additional hyperparameter tuning. 
	
	The approach works well with synthetically generated events from an existing dense dataset, which we believe will be beneficial for the adoption of this technology. Whereas the production of new high-quality datasets for specialised application domains can be very expensive, dense datasets are quite abundant in comparison, even with the constraint of requiring image-sequences to be applicable for conversion to events.
	
	We think that asynchronous-sparse \glspl{cnn} are a promising new concept that may find use in real-time applications due to the extremely low latency.
	In practice, however, these concepts are not yet sufficiently optimized.
		
	We have extended the experimental asynet framework for asynchronous-sparse \glspl{cnn} and shown that sparse \glspl{cnn} match classical dense \glspl{cnn} in prediction performance.
	On the other hand, we found that the runtime performance of the evaluated frameworks cannot yet match dense networks, and especially the asynchronous framework can at this point only be seen as a proof-of-concept.
	The ease of transfer from dense to sparse networks and the potential gains in runtime will hopefully incite further research into these promising technologies.
    We believe the evaluated sparse \gls{cnn} frameworks to be limited by code inefficiencies and lacking hardware support, but in theory to be a viable optimization of \glspl{cnn}.

\bibliographystyle{ACM-Reference-Format}
\bibliography{bibliography}


\begin{thebibliography}{10}


\ifx \showCODEN    \undefined \def \showCODEN     #1{\unskip}     \fi
\ifx \showDOI      \undefined \def \showDOI       #1{#1}\fi
\ifx \showISBNx    \undefined \def \showISBNx     #1{\unskip}     \fi
\ifx \showISBNxiii \undefined \def \showISBNxiii  #1{\unskip}     \fi
\ifx \showISSN     \undefined \def \showISSN      #1{\unskip}     \fi
\ifx \showLCCN     \undefined \def \showLCCN      #1{\unskip}     \fi
\ifx \shownote     \undefined \def \shownote      #1{#1}          \fi
\ifx \showarticletitle \undefined \def \showarticletitle #1{#1}   \fi
\ifx \showURL      \undefined \def \showURL       {\relax}        \fi
\providecommand\bibfield[2]{#2}
\providecommand\bibinfo[2]{#2}
\providecommand\natexlab[1]{#1}
\providecommand\showeprint[2][]{arXiv:#2}

\bibitem[\protect\citeauthoryear{Cannici, Ciccone, Romanoni, and
  Matteucci}{Cannici et~al\mbox{.}}{2019}]%
        {yole}
\bibfield{author}{\bibinfo{person}{Marco Cannici}, \bibinfo{person}{Marco
  Ciccone}, \bibinfo{person}{Andrea Romanoni}, {and} \bibinfo{person}{Matteo
  Matteucci}.} \bibinfo{year}{2019}\natexlab{}.
\newblock \showarticletitle{Asynchronous convolutional networks for object
  detection in neuromorphic cameras}. In \bibinfo{booktitle}{\emph{Proceedings
  of the IEEE/CVF Conference on Computer Vision and Pattern Recognition
  Workshops}}.
\newblock


\bibitem[\protect\citeauthoryear{Gallego, Delbruck, Orchard, Bartolozzi, Taba,
  Censi, Leutenegger, Davison, Conradt, Daniilidis, and Scaramuzza}{Gallego
  et~al\mbox{.}}{2020}]%
        {event-vision}
\bibfield{author}{\bibinfo{person}{Guillermo Gallego}, \bibinfo{person}{Tobi
  Delbruck}, \bibinfo{person}{Garrick~Michael Orchard}, \bibinfo{person}{Chiara
  Bartolozzi}, \bibinfo{person}{Brian Taba}, \bibinfo{person}{Andrea Censi},
  \bibinfo{person}{Stefan Leutenegger}, \bibinfo{person}{Andrew Davison},
  \bibinfo{person}{J{\"o}rg Conradt}, \bibinfo{person}{Kostas Daniilidis},
  {and} \bibinfo{person}{Davide Scaramuzza}.} \bibinfo{year}{2020}\natexlab{}.
\newblock \showarticletitle{Event-based Vision: A Survey}.
\newblock \bibinfo{journal}{\emph{IEEE Transactions on Pattern Analysis and
  Machine Intelligence}} (\bibinfo{year}{2020}).
\newblock
\showISSN{1939-3539}
\urldef\tempurl%
\url{https://doi.org/10.1109/tpami.2020.3008413}
\showDOI{\tempurl}


\bibitem[\protect\citeauthoryear{Gehrig, Gehrig, Hidalgo-Carri\'o, and
  Scaramuzza}{Gehrig et~al\mbox{.}}{2020}]%
        {vid2e}
\bibfield{author}{\bibinfo{person}{Daniel Gehrig}, \bibinfo{person}{Mathias
  Gehrig}, \bibinfo{person}{Javier Hidalgo-Carri\'o}, {and}
  \bibinfo{person}{Davide Scaramuzza}.} \bibinfo{year}{2020}\natexlab{}.
\newblock \showarticletitle{Video to Events: Recycling Video Datasets for Event
  Cameras}. In \bibinfo{booktitle}{\emph{{IEEE} Conf. Comput. Vis. Pattern
  Recog. (CVPR)}}.
\newblock


\bibitem[\protect\citeauthoryear{Geiger}{Geiger}{2017}]%
        {kitti-web}
\bibfield{author}{\bibinfo{person}{Andreas Geiger}.}
  \bibinfo{year}{2017}\natexlab{}.
\newblock \bibinfo{booktitle}{\emph{The KITTI Vision Benchmark Suite}}.
\newblock
\urldef\tempurl%
\url{http://www.cvlibs.net/datasets/kitti/eval_object.php}
\showURL{%
\tempurl}


\bibitem[\protect\citeauthoryear{Geiger, Lenz, and Urtasun}{Geiger
  et~al\mbox{.}}{2012}]%
        {kitti}
\bibfield{author}{\bibinfo{person}{Andreas Geiger}, \bibinfo{person}{Philip
  Lenz}, {and} \bibinfo{person}{Raquel Urtasun}.}
  \bibinfo{year}{2012}\natexlab{}.
\newblock \showarticletitle{Are we ready for Autonomous Driving? The KITTI
  Vision Benchmark Suite}. In \bibinfo{booktitle}{\emph{Conference on Computer
  Vision and Pattern Recognition (CVPR)}}.
\newblock


\bibitem[\protect\citeauthoryear{Graham, Engelcke, and van~der Maaten}{Graham
  et~al\mbox{.}}{2018}]%
        {scn}
\bibfield{author}{\bibinfo{person}{Benjamin Graham}, \bibinfo{person}{Martin
  Engelcke}, {and} \bibinfo{person}{Laurens van~der Maaten}.}
  \bibinfo{year}{2018}\natexlab{}.
\newblock \showarticletitle{3D Semantic Segmentation with Submanifold Sparse
  Convolutional Networks}.
\newblock \bibinfo{journal}{\emph{CVPR}} (\bibinfo{year}{2018}).
\newblock


\bibitem[\protect\citeauthoryear{Maass}{Maass}{1997}]%
        {snn}
\bibfield{author}{\bibinfo{person}{Wolfgang Maass}.}
  \bibinfo{year}{1997}\natexlab{}.
\newblock \showarticletitle{Networks of spiking neurons: the third generation
  of neural network models}.
\newblock \bibinfo{journal}{\emph{Neural networks}} \bibinfo{volume}{10},
  \bibinfo{number}{9} (\bibinfo{year}{1997}), \bibinfo{pages}{1659--1671}.
\newblock


\bibitem[\protect\citeauthoryear{Messikommer, Gehrig, Loquercio, and
  Scaramuzza}{Messikommer et~al\mbox{.}}{2020}]%
        {rpg}
\bibfield{author}{\bibinfo{person}{Nico Messikommer}, \bibinfo{person}{Daniel
  Gehrig}, \bibinfo{person}{Antonio Loquercio}, {and} \bibinfo{person}{Davide
  Scaramuzza}.} \bibinfo{year}{2020}\natexlab{}.
\newblock \showarticletitle{Event-based Asynchronous Sparse Convolutional
  Networks}.
\newblock \bibinfo{journal}{\emph{European Conference on Computer Vision.
  (ECCV)}}.
\newblock
\urldef\tempurl%
\url{http://rpg.ifi.uzh.ch/docs/ECCV20_Messikommer.pdf}
\showURL{%
\tempurl}


\bibitem[\protect\citeauthoryear{Rebecq, Ranftl, Koltun, and Scaramuzza}{Rebecq
  et~al\mbox{.}}{2019}]%
        {e2vid}
\bibfield{author}{\bibinfo{person}{Henri Rebecq}, \bibinfo{person}{Ren{\'{e}}
  Ranftl}, \bibinfo{person}{Vladlen Koltun}, {and} \bibinfo{person}{Davide
  Scaramuzza}.} \bibinfo{year}{2019}\natexlab{}.
\newblock \showarticletitle{Events-to-Video: Bringing Modern Computer Vision to
  Event Cameras}.
\newblock \bibinfo{journal}{\emph{{IEEE} Conf. Comput. Vis. Pattern Recog.
  (CVPR)}} (\bibinfo{year}{2019}).
\newblock


\bibitem[\protect\citeauthoryear{Redmon, Divvala, Girshick, and Farhadi}{Redmon
  et~al\mbox{.}}{2016}]%
        {yolo-paper}
\bibfield{author}{\bibinfo{person}{Joseph Redmon}, \bibinfo{person}{Santosh
  Divvala}, \bibinfo{person}{Ross Girshick}, {and} \bibinfo{person}{Ali
  Farhadi}.} \bibinfo{year}{2016}\natexlab{}.
\newblock \showarticletitle{You only look once: Unified, real-time object
  detection}. In \bibinfo{booktitle}{\emph{Proceedings of the IEEE conference
  on computer vision and pattern recognition}}. \bibinfo{pages}{779--788}.
\newblock


\end{thebibliography}


\end{document}